\documentclass[11pt,a4paper]{article}
\usepackage[hyperref]{naaclhlt2019}
\usepackage{times}
\usepackage{latexsym}

\usepackage{url}

\usepackage{pifont}
\usepackage{helvet}
\usepackage{courier}
\usepackage{tabularx}
\usepackage{import}
\usepackage{graphicx}
\usepackage{color}
\usepackage{array}
\usepackage[normalem]{ulem} 
\usepackage{cancel} 
\usepackage{booktabs} 
\usepackage[title]{appendix} 
\usepackage{enumitem} 

\newcommand{\eg}{e.g.,}
\newcommand{\ie}{i.e.,}
\newcommand{\etc}{etc.}

\definecolor{darkGreen}{rgb}{0,0.6,0}

\newcolumntype{L}[1]{>{\raggedright\let\newline\\\arraybackslash\hspace{0pt}}m{#1}}
\newcolumntype{C}[1]{>{\centering\let\newline\\\arraybackslash\hspace{0pt}}m{#1}}
\newcolumntype{R}[1]{>{\raggedleft\let\newline\\\arraybackslash\hspace{0pt}}m{#1}}

\aclfinalcopy 
\def\aclpaperid{1498} 

\newcommand{\descrcell}[2]{%
  \scriptsize
  \begin{tabular}[t]{@{}c@{}}\normalsize#1\\#2\end{tabular}%
}

\title{``President Vows to Cut \sout{Taxes} \underline{Hair}'':\\ Dataset and Analysis of Creative Text Editing for Humorous Headlines}

\author{Nabil Hossain\hspace{35pt} \\
  Dept. Computer Science\hspace{35pt} \\
  University of Rochester\hspace{35pt} \\
  Rochester, NY\hspace{35pt} \\
  {\tt nhossain@cs.rochester.edu}\hspace{35pt} \And 
  \hspace{20pt}John Krumm \and Michael Gamon\\
  \hspace{20pt}Microsoft Research AI \\
  \hspace{20pt}Microsoft Corporation \\
  \hspace{20pt}Redmond, WA \\
  \hspace{20pt}{\tt \{jckrumm,mgamon\}@microsoft.com}} 

\date{}

\begin{document}
\maketitle


\begin{abstract}
We introduce, release, and analyze a new dataset, called Humicroedit, for research in computational humor. Our publicly available data consists of regular English news headlines paired with versions of the same headlines that contain simple replacement edits designed to make them funny. We carefully curated crowdsourced editors to create funny headlines and judges to score a to a total of 15,095 edited headlines, with five judges per headline. The simple edits, usually just a single word replacement, mean we can apply straightforward analysis techniques to determine what makes our edited headlines humorous. We show how the data support classic theories of humor, such as incongruity, superiority, and setup/punchline. Finally, we develop baseline classifiers that can predict whether or not an edited headline is funny, which is a first step toward automatically generating humorous headlines as an approach to creating topical humor.
\end{abstract}
\section{Introduction}
Humor detection and generation continue to be challenging AI problems.  
While there have been some advances in automatic humor recognition ~\cite{khodak2017large,davidov2010semi,barbieri2014automatic,reyes2012humor,cattle2018recognizing,bertero2016long,yang2015humor}, computerized humor generation has seen less progress \cite{binsted1997children,stock2003hahacronym,petrovic2013unsupervised}.
This is not surprising, given that humor involves in-depth world-knowledge, common sense, and the ability to perceive relationships across entities and objects at various levels of understanding. Even humans often fail at being funny or recognizing humor. 

A big hindrance to progress on humor research is the scarcity of public datasets. Furthermore, the existing datasets address specific humor templates,
such as funny one-liners~\cite{mihalcea2006learning} and filling in Mad Libs\textsuperscript{\small{\textregistered}}~\cite{hossain2017filling}. Creating a humor corpus is non-trivial, however, because it requires (i) human annotation, and (ii) a clear definition of humor to achieve good inter-annotator agreement.

\begin{figure}
    \centering
    \begin{tabular}{c}
      \resizebox{0.95\columnwidth}{!}{\fbox{
        \includegraphics{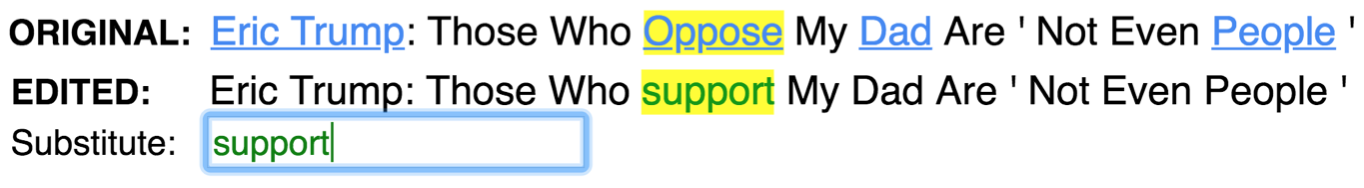}}} \\ 
        \descrcell{\small{(a) The  Headline Editing Task. }}{\\} \\
      \resizebox{0.95\columnwidth}{!}{\fbox{
        \includegraphics{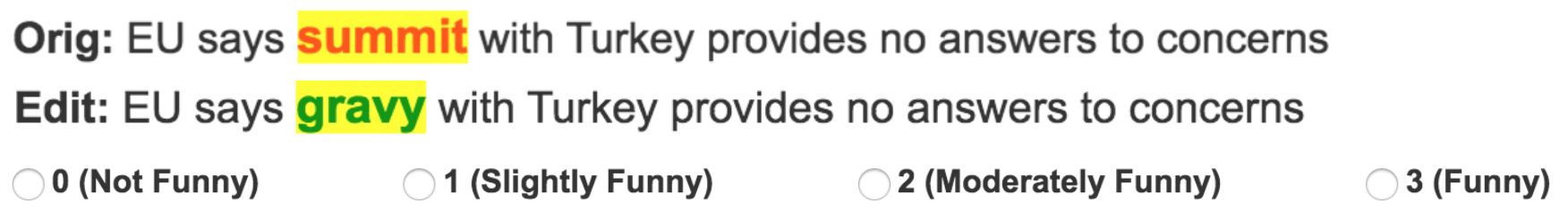}}} \\ \small {(b) The Headline Grading Task.} \\
    \end{tabular}
  \caption{Snapshots of the headline editing and grading interfaces. Only the underlined tokens are replaceable.}
  \label{fig:mturkScreenShots}
\end{figure}

We introduce {\bf Humicroedit}, a novel dataset for research in computational humor. First, we collect original news headlines from news media posted on Reddit (\texttt{reddit.com}). Then, we {\it qualify} expert annotators from Amazon Mechanical Turk (\texttt{mturk.com}) to (i) generate humor by applying small edits to these headlines, and to (ii) judge the humor in these edits. Our resulting dataset contains 15,095 edited news headlines and their numerically assessed humor. Screenshots of our two annotation tasks are shown in Figure~\ref{fig:mturkScreenShots}, and Table~\ref{tab:headline_edits} shows some of these annotated headlines. 

This new dataset enables various humor tasks, such as: (i) understanding what makes an edited headline funny,
(ii) predicting whether an edited headline is funny, (iii) ranking multiple edits of the same headline on a funniness scale, (iv) generating humorous news headlines, and (v) recommending funny headlines personalized to a reader.

Our dataset presents several opportunities for computational humor research since: 
\begin{itemize}[leftmargin=*]\itemsep-0.2em 
    \item Headlines do not have specific templates.
    \item Headlines contain very few words, but convey a lot of information. 
    \item A deeper understanding of world-knowledge and common-sense is needed to completely understand what makes a headline funny.
    \item Humorous headlines are often generated using several layers
    of cognition and reasoning.
    \item Despite us carefully qualifying annotators, their knowledge, preferences, bias and stance towards 
    information presented in headlines influence whether they perceive a potentially funny headline as humorous, offensive, confusing, \etc{}
\end{itemize}
The presence of these factors suggests that thorough humor comprehension in our dataset requires the development of NLP tools that are not only robust at pattern recognition but also capable of deeper semantic understanding and reasoning. 
As an initial exploration of this proposition, we perform various data analysis against the background of humor theories, and we train and examine classifiers to detect humorous edited headlines in our data.

 \begin{table*}
     \centering
     \resizebox{0.99\textwidth}{!}{        
    \begin{tabular}{|c|l|l|c|c|}
        \hline
         {\bf ID} & {\bf Original Headline} & {\bf Substitute} & {\bf Grade} & {\bf Prediction}\\ \hline
          1 & Kushner to visit {\bf \textcolor{red}{Mexico}} following latest Trump tirades & \textcolor{darkGreen}{therapist} & 2.8 & \ding{56} \\ \hline 
          2 & Trump wants you to take his {\bf \textcolor{red}{tweets}} seriously. His aides don't & \textcolor{darkGreen}{hair} & 2.8 & \ding{52} \\ \hline
          3 & Essential Politics: California's hottest congressional {\bf \textcolor{red}{races}}, ranked & \textcolor{darkGreen}{mistresses} & 2.8 & \ding{56} \\ \hline
          4 & Hillary Clinton Staffers Considered Campaign Slogan `Because It's Her {\bf \textcolor{red}{Turn}}'  & \textcolor{darkGreen}{fault} & 2.8 & \ding{56} \\ \hline

          5 & Trump Vows North Korea Could be Met With `Fire and {\bf \textcolor{red}{Fury}}' & \textcolor{darkGreen}{marshmallows} & 2.6 &  \ding{52} \\ \hline
          
          6 & Here's how {\bf \textcolor{red}{Wall}} Street is reacting to Trump's tax plan &	\textcolor{darkGreen}{sesame} & 2.4 & \ding{52} \\ \hline
          
        7 & Swedish prosecutor says {\bf \textcolor{red}{truck}} attack suspect has not spoken &	\textcolor{darkGreen}{mime} & 2.4 & \ding{56}  \\ \hline
        
        8 & {\bf \textcolor{red}{Steve Bannon}} questioned by special counsel	& \textcolor{darkGreen}{kindergarteners} & 2.4 &   \ding{56}  \\ \hline
        
          9 & New survey shows majority of US troops has `unfavorable' view of Obama's {\bf \textcolor{red}{years}} & \textcolor{darkGreen}{ears} & 2.2 & \ding{56} \\ \hline

          10 & The Latest: BBC cuts {\bf \textcolor{red}{ties}} with Myanmar TV station  & \textcolor{darkGreen}{pies} & 1.8 & N/A (train) \\ \hline
          
         11 & Bill Maher: ``I {\bf \textcolor{red}{doubt}} that Trump will be president the full term'' & \textcolor{darkGreen}{hope} & 0.2 & \ding{56}  \\ \hline

          12 & Malawi arrests 140 in clampdown after `vampirism' {\bf \textcolor{red}{killings}}	&
        \textcolor{darkGreen}{rumors} & 0.2 & \ding{56}  \\ \hline

         13 & {\bf \textcolor{red}{Rising}} Dem star announces engagement to same-sex partner	& \textcolor{darkGreen}{gay} & 0.0 & \ding{56}  \\ \hline
				
        14 & 	{\bf \textcolor{red}{Taylor Swift}} claims Denver DJ sexually assaulted her back in 2013	& \textcolor{darkGreen}{hen} & 0.0 & \ding{56}  \\ \hline
        
        15 & 4 {\bf \textcolor{red}{soldiers}} killed in Nagorno-Karabakh fighting: Officials
    	& \textcolor{darkGreen}{rabbits} & 0.0 & \ding{52}  \\ \hline
        
        \end{tabular}}
    \vspace{-4pt}
    \caption{Some headlines in our dataset and their edits, mean funniness grades, and accuracy of funniness prediction by LSTM. 
     We acknowledge that some of these
     are offensive, but we use them for analysis in Section 4.1.}
     \label{tab:headline_edits}
    \vspace{-2pt}
 \end{table*}

\section{The Humor Dataset}
In this section, we describe how we gathered our set of original headlines, directed editors to make them funny, employed graders to assess the level of humor in the modified headlines, and created the Humicroedit dataset.

\subsection{Task Description}
Our goal is to study how humor is generated by applying short edits to headlines. News headlines are ripe for humor, since they convey rich information using only a few words. While the short form may seem to limit context, readers have rich background information in the form of their existing world knowledge, which helps them understand the headline. 
Allowing only short edits means we can apply focused analysis on the tipping point between regular and funny.

Therefore, our task is to edit a headline to make it funny, where an edit 
is defined as the insertion of a single-word noun or verb to replace an existing entity or single-word noun or verb.
Note that our rules do not allow:
\begin{itemize}[leftmargin=*]
\itemsep-0.2em 
    \item Addition/removal of a whole noun/verb phrase, except removal of noun phrases that are entities (\eg{} One World, Virtual Reality).
    \item Removal of sub-tokens within entities (\eg{} replacing only ``States'' in ``United States'').
\end{itemize}
The decision to strictly avoid edits of other parts-of-speech (POS) words was motivated by the observation in our pilot experiments that those edits did not provide enough variety of humor. For example, when substituting adjectives and adverbs, our editors mostly used antonyms or superlatives.
Switching nouns and verbs, on the other hand, enables the introduction of diverse novel connections between entities and actions.

To identify the replaceable entities, we apply named entity recognition (NER) and POS tagging using the Stanford CoreNLP toolkit~\cite{manning-corenlp}. We allow for replacement of only those entities that are well-known, according to the Microsoft Knowledge Base\footnote{http://blogs.bing.com/search/2013/03/21/understand-your-world-with-bing/}. This improves the likelihood that the terms are familiar to both headline editors and humor judges. 
We allow a noun (or verb) to be replaced if it is an unambiguous noun (or verb) in WordNet~\cite{fellbaum1998wordnet} (\ie{} has a single WordNet POS). Editors are only allowed to replace one of the selected replaceable words/entities in the headline.

We refer to a single-term substitution of this type as a ``{\bf micro-edit}'', and we will use this term interchangeably with ``edit'' in the remainder of this paper. Micro-edits approach the smallest change that can induce humor in text, letting us focus intently on what causes humor.

\subsection{Collecting Headlines}
We build our dataset from popular news headlines posted on the social media site Reddit. 
This strategy steers us towards a set of headlines that is  part of general discourse, rather than being only of specialized interest, which would make editing them for humor difficult. 

We obtain all Reddit posts from the popular subreddits \texttt{r/worldnews} and \texttt{r/politics} 
from January 2017 to May 2018 using Google BigQuery\footnote{\texttt{https://cloud.google.com/bigquery}}. Each of these posts is a headline from a news source. 
We remove duplicate headlines and headlines that have fewer than 4 words or more than 20 words. 
Finally, we keep only the headlines from the 25 English news sources that contribute the most headlines in the Reddit data, resulting in a total of 287,076 news  headlines.

\subsection{Annotation}
For our data annotation tasks, we use Mechanical Turk workers who (i) are located in the U.S., (ii) have a HIT approval rate greater than 97\%, and (iii) have more than 10,000 HITs approved.
To ensure high data quality, we further qualify distinct sets of (i) turker judges for recognizing humor in an edited headline, and (ii) editors adept at editing headlines to generate humor. 
\subsubsection{Qualifying Humor Judges}
We manually collected a set of 20 original news headlines and edited each of them such that some edits are funny and some are not. We asked 
several members of our research group to assess the funniness of each edited headline using the following integer scale developed by \citet{hossain2017filling}:
\begin{center}
\begin{tabular}{lcl}
{\bf 0} - Not funny   &  &  {\bf 1} - Slightly funny \\
{\bf 2} - Moderately funny &   & {\bf 3} - Funny
\end{tabular}
\end{center}
\vspace{1pt}
We instructed internal and turker judges (i) to grade objectively regardless of their own stance towards issues, entities and information expressed in the headline, and (ii) to grade an edited headline as funny if they believed it would be funny to a large audience.
Further, we instructed judges to grade an edited headline as funny if either the headline was funny by itself regardless of the original headline, or the headline was only funny when considering how the original headline was changed.
    
We labeled the ground truth funniness of each of these 20 edited qualifier headlines as its mean internal judge grade. For the qualification task, we classified as funny any edited headline with a mean grade of 1.0 or above.

Next, we launched the same task on Mechanical Turk until we found 150 qualified judges (60\% of the candidates). Turkers were qualified if (i) they had 3 or fewer classification errors according to our 1.0 threshold, and (ii) on average, their grades were within 0.6 of the mean internal judge grades.

We calculated the inter-annotator agreement for assigning headline funniness grades using the Krippendorff's $\alpha$ interval metric~\cite{krippendorff1970estimating}  --- a real number in the range $[-1,1]$, with -1, 0 and 1, respectively, implying complete disagreement, no consensus and full agreement. 
The $\alpha$ for the internal judges and qualified turker judges were, respectively, 0.57 and 0.64. 

\subsubsection{Qualifying Humor Editors}
For editor qualification, we randomly sampled 60 headlines, split into 6 separate Mechanical Turk tasks of 10 headlines each. Candidate editors were asked to complete one of these tasks, which was to make each headline as funny as possible to a general audience using a micro-edit. Task participants were instructed not to apply the following edits:
\begin{itemize}[leftmargin=*]
\itemsep-0.2em 
    \item Cheap humor generation techniques: add profanity, slang, bathroom/potty humor, \emph{crude} sexual references or informal language.
    \item Squeeze multiple words into one (\eg{} Housecat, JumpedOverWall).
\end{itemize}
Next, we used 7 qualified judges to assess the funniness of each edited headline of each candidate. We qualified all candidates whose mean funniness of edited headlines was above 0.8 or the task's average headline's funniness grade, whichever was higher. In total, we obtained 100 qualified editors (57.5\% of the candidates) who met our expectations in their ability to create funny headlines.

\subsection{Data Collection and Quality Control}
For our final dataset, we randomly sampled a total of 5,170 news headlines from our Reddit dataset, obtaining roughly an equal number of headlines from each news source. We asked 3 editors to edit each headline and 5 judges to grade each edited headline. Multiple micro-edits of the same headline allow us to compare different edits in terms of their effectiveness for generating humor, which we leave for future work.

To avoid turker exhaustion and decision fatigue, we performed the annotation task over a series of mini-batches launched at least 24 hours apart. After each round of editing, we applied tools to (i) check the edits for spelling mistakes which we manually corrected, and (ii) to find and eliminate inserted tokens that were a concatenation of two or more words (\eg{} selftanner). To allow diversity in annotations, we applied a maximum HIT limit for annotators per batch. After each batch was completed, we temporarily suspended those editors and judges who had done significantly more HITs than the rest, until the others caught up. 

Lastly, as we obtained more and more annotated data, the editors started employing the same humor generation strategies (\eg{} inserting words from a small vocabulary). Consequently, judges saw repeated, identical edits, so the element of surprise was gone, and the judges were grading fewer humorous edited headlines as funny.
We addressed this by randomly sampling a set of editors and judges for each batch, obtaining new editors and judges over time, and removing those editors who had done a majority of the HITs but whose edits' average funniness grade fell below a threshold (=0.7) after they participated in a batch. 
We also removed judges who repeatedly assigned very low funniness grades compared to the 4 other judges for the same edit. The judges' agreement score based on $\alpha$ was 0.20,  showing modest agreement considering the factors above and others such as judges' personal preferences, bias, political stance, \etc{} which make consensus difficult.

Our Humicroedit dataset includes 15,095 unique edited headlines
graded for funniness.
For annotating a single headline, we paid 10 US cents to editors and 2.5 US cents to judges. 
There were also small costs for qualification. Our total cost for obtaining the dataset is about USD 4,500\footnote{Dataset: \href{www.cs.rochester.edu/u/nhossain/humicroedit.html}{cs.rochester.edu/u/nhossain/humicroedit.html}.

Total cost is USD 4,500, not USD $4,500^{\bf 3}$ (joke!)
}.

\section{Humor Analysis}
In this section, we analyze what types of micro-edits are effective at creating humor in our dataset, and we discuss our findings against the background of humor theories.

\begin{figure}
    \centering
    \resizebox{\columnwidth}{!}{
        \includegraphics{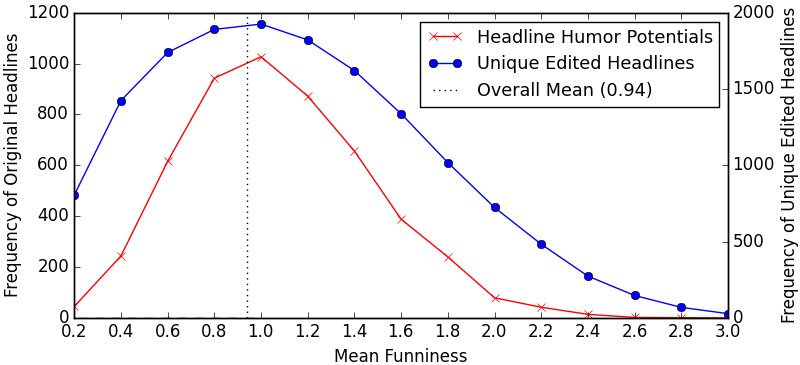}}
    \caption{Histograms of humor potentials of headlines and mean funniness of unique edited headlines. The humor potential of a headline is the mean funniness score over all edits, shown as the red curve. The blue curve shows the histogram of the mean score of each distinct edited headline.}
    \label{fig:headline_funniness}
\end{figure}

Figure~\ref{fig:headline_funniness} shows the histogram of the mean rating of each edited headline. While the majority of the headlines achieve slight to moderate levels of humor, some of them appear inherently difficult to make humorous by micro-editing.
We noticed that editors encountered difficulty making headlines funny when the headlines had very negative themes, such as shootings, death, \etc{}, and when they focused on information less likely to be known by a general audience (\eg{} relatively unknown  person, an insignificant political issue).

\subsection{Humor Generation Strategies}
By manual inspection, we can gain insights into humor generation strategies employed by our editors, which we discuss with references to Table~\ref{tab:headline_edits}:
\begin{enumerate}[leftmargin=*]
\itemsep-0.2em 
    \item Using a word that forms a meaningful n-gram with the adjacent words (\eg{} ID 5: Fire and \sout{Fury} \underline{marshmallows}; ID 6: \sout{Wall} \underline{sesame} street).
    \item Connection between replaced word and the replacement: replacements that are semantically distant from (\eg{} ID 1: \sout{Mexico} \underline{therapist}) or similar in pronunciation to (\eg{} ID 10: \sout{ties} \underline{pies}) to the replaced word.
    \item Using a word that makes a strong connection with an entity in the headline (\eg{} ID 2:  Trump and hair; ID 9: Obama and ears).
    \item Creating sarcasm (ID: 11).
    \item Belittling an entity or noun in the headline (\eg{} ID 4: Hillary Clinton's \sout{turn} \underline{fault}; ID 9: Obama's \sout{years} \underline{ears}).
    \item Tension suppression\footnote{This is also known as the relief theory of humor.}: making a serious headline silly (\eg{} IDs 5 and 9).
    \item Inserting words that generate incongruity (common among most examples in Table~\ref{tab:headline_edits}).
    \item Setup and punchline: let the headline build up towards an expected ending, and then change words towards the end to produce a coherent but surprising ending (\eg{} IDs 3, 4 and 5).
\end{enumerate}

\subsection{Clusters of Replacement Words}
Each micro-edit used a new replacement word to change the headline. We clustered these replacement words using their GloVe word vectors~\cite{pennington2014glove} and $k$-means clustering, with $k=20$. Our manually-generated cluster names are shown in Table \ref{tab:nabil_clusters}, where the clusters are ordered by the mean funniness score of the edited headlines whose replacement word is in the cluster. For each cluster, we show the frequency with which the cluster was used for replacement words and frequent sample words from the cluster.

We can compare our automatically generated clusters with those of~\citet{westbury2018wriggly}. They manually created six clusters from the 200 funniest, single words in~\citet{engelthaler2018humor} and then they added more words algorithmically. Four of their six manually curated classes have direct correspondences to our automatically curated classes: \emph{sex}, \emph{insults}, \emph{bodily functions}, and \emph{animals}. We did not find an equivalent to their \emph{profanity} class, because we instructed our editors to avoid profanity. There is also a \emph{party} class that we do not have. Overall, though, we find good agreement between their manually curated classes and some of our automatically generated clusters, leading us to believe that our clusters are meaningfully representative of humor generation strategies for our task.

\begin{table*}
\begin{center}
\resizebox{0.935\textwidth}{!}{
\begin{tabular}{@{}lllll@{}}
\toprule
Class Label                   & Funniness & Frequency \%     & 5 Frequent Sample Words                                    &  \\ \midrule
clothes and fashion           & 1.213           & 5.77      & hair, pants, haircut, fashion, underwear        &  \\
sex                           & 1.209           & 3.52      & orgy, spank, mistress, porn, striptease         &  \\
food (savory)                 & 1.113           & 3.89      & cheese, sandwich, chicken, potato, tacos        &  \\
eating                        & 1.048           & 4.18      & food, pizza, dinner, restaurant, eat            &  \\
fantasy characters            & 1.036           & 4.94      & clown, aliens, penguin, robot, ghost            &  \\
music and shows               & 1.023           & 3.85      & dance, circus, music, sings, mime               &  \\
bodily functions              & 1.021           & 3.11      & diet, brain, odor, dandruff, pimple             &  \\
food (snacks)                 & 1.017           & 5.36      & pumpkin, cake, vodka, cookies, candy            &  \\
animals                       & 1.000           & 9.22      & dog, monkey, puppy, cats, duck                  &  \\
various nouns (1)             & 0.935           & 5.82      & toupee, hoedown, jaywalking, barbers, seance    &  \\
jobs and roles                & 0.890           & 4.95      & children, wife, baby, mother, barber            &  \\
insults                       & 0.876           & 3.36      & clowns, tantrum, trolls, racist, whining        &  \\
emotional                     & 0.868           & 7.74      & love, hug, jokes, fights, cry                   &  \\
leisure             & 0.862           & 6.01      & vacation, shopping, tanning, hotel, pool       &  \\
sports                        & 0.856           & 3.03      & horse, game, bowling, golf, wrestling           &  \\
various nouns (2)             & 0.850           & 4.90      & water, nose, balloon, gas, smoke                &  \\
human deficiencies            & 0.803           & 4.21      & lies, ignorance, humor, boredom, stupidity      &  \\
media                         & 0.787           & 4.20      & tweet, movie, book, video, television           &  \\
aspirations                   & 0.726           & 7.28      & party, date, people, money, model               &  \\
corrupted                      & 0.712           & 4.48      & president, bribes, politicians, destroy, prison &  \\ \bottomrule
\end{tabular}
}
\end{center}
\vspace{-8pt}
\caption{Twenty clusters of replacement words with manually determined cluster labels.}
\label{tab:nabil_clusters}
\end{table*}

\subsection{Support for Theories of Humor}
Our rated headlines give us an opportunity to explore theories of humor in a systematic way. We find, in general, that these theories are supported by our data.

\subsubsection{Length of Joke}
Although some linguists argue that jokes should make economical use of words~\cite{tomoioagua2015linguistic},~\citet{ritchie2004linguistic} argues that jokes often have extra information, which can make a joke funnier. While humorous headlines form a special niche of jokes, we observed that longer headlines generally had higher humor potential.

\begin{figure}
    \centering
    \includegraphics[width=\columnwidth]{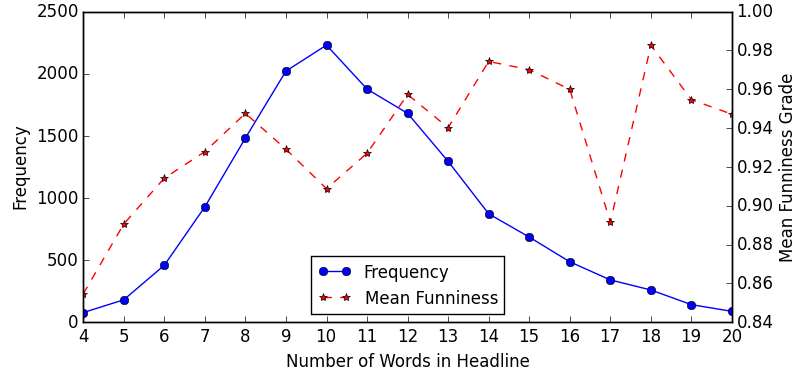}
    \caption{Short headlines did not lend themselves to high humor scores, while longer headlines generally had more potential for humor. The blue line shows the raw distribution of headline lengths in our data, and the red line shows the mean funniness score over different lengths.}
    \label{fig:headline_vs_words}
\end{figure}

Figure~\ref{fig:headline_vs_words} shows that the population of our collected headlines from Reddit has a length distribution with a peak at 10 words and a long tail to the right. The least funny edited headlines are the shortest, and the most funny are the longest. This makes sense since very short headlines (4-5 words long) barely have enough contextual information to exploit to make a humorous edit, whereas headlines that have very rich contexts generally allow editors more flexibility to generate humor. We note that~\citet{dunbar2016complexity} also found that longer jokes are funnier, but that some jokes could be too complicated to be funny.

We can also examine the number and proportion of \emph{replaceable} words and how these numbers affect funniness. In our dataset, the number of replaceable words ranged between 1 and 12, and funniness grades of micro-edits were significantly lower at the two extremes. Editors apparently had difficulty generating humor when they were severely constrained in choosing a word to replace, or when they had too many choices for replacement. However, edited headlines with a higher proportion of replaceable words were generally funnier, as shown in Table~\ref{tab:funny_vs_proportion}.
This suggests that allowing editors more freedom in choosing words from the headline to edit results in better humor, or that high proportion of nouns, entities and verbs in the headline increases the chance of successful humor generation.

\begin{table}
    \centering
  \resizebox{\columnwidth}{!}{
    \begin{tabular}{|l||c|c|c|c|c|}
    \hline
         Repl. Words Prop. & 0.2 & 0.4 & 0.6 & 0.8 & 1.0 \\ \hline
         Mean Funniness & 0.87 & 0.92 & 0.95 & 0.94 & 1.0 \\ \hline
    \end{tabular}
    }
    \caption{Edited headlines were judged funnier when they had a larger proportion of replaceable words.}
    \label{tab:funny_vs_proportion}
\end{table}

\subsubsection{Incongruity for Humor}
We see evidence for the incongruity theory of humor~\cite{sep-humor}. Jokes that use incongruity aim to violate an expectation, with the expectation normally set up by the joke itself. We test incongruity by examining the relationship between the replacement words chosen by our editors and words in the original headline using cosine distances between their GloVe vectors.
If incongruity is important, we expect the replacement word to be distant from the headline's original words.

\begin{figure}
    \centering
    \includegraphics[width=\columnwidth]{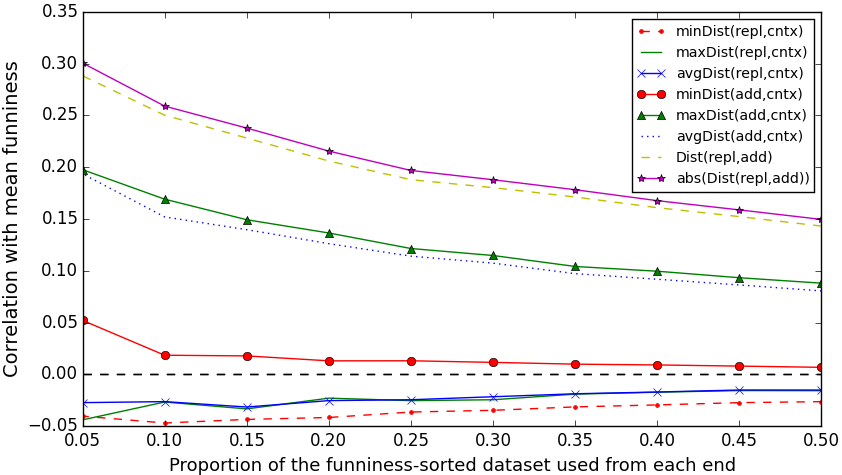}
    \caption{Correlations of word vector based cosine distances with mean funniness at various dataset sizes. We measure distance (and its absolute value) between the replaced word(s) and the added word, and also their minimum, maximum and average distances with the set of other words in the headline.}
    \label{fig:gloveCorrelations}
\end{figure}

The results of this analysis are shown in Figure~\ref{fig:gloveCorrelations}. Our approach involved computing the correlations between mean funniness scores of edited headlines and different GloVe distances between their replacement words and the other words in the headline serving as context. In order to sharpen the analysis, we looked at subsets of headlines with extreme funniness scores. For instance, the left-most data points in Figure~\ref{fig:gloveCorrelations} pertain only to those edited headlines that are in the top and bottom 5\% of mean scores, which filters out headlines whose scores are in the middle. 

The four curves higher on the plot show a relatively high correlation between humor scores and the cosine distance between the added word (``add'' in legend) and the replaced word (``repl'' in legend) or the other words in the headline (``cntx'' in legend). This suggests that incongruity leads to humor. The three lower curves show there is not a strong correlation between humor and the distance between the original, replaced word the other words in the headline. Finally, smaller, less humor-ambiguous data leads to stronger positive correlations, which suggests that higher incongruity leads to more quality humor.

\subsubsection{Setup and Punchline}
\begin{figure}
    \centering
    \includegraphics[width=\columnwidth]{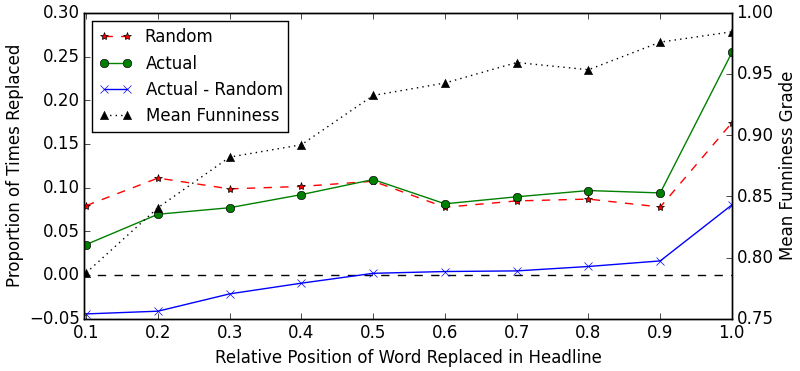}
    \caption{The joke profile graph for setup and punchline humor, showing that word substitutions toward the end of the headline normally lead to better humor.}
    \label{fig:joke_profile}
\end{figure}

We specifically studied whether the ``setup and punchline''~\cite{rochmawati2017pragmatic} approach is used in funny headline generation, where the humor comes toward the end of the joke after a setup at the beginning. This has been verified numerically for funny cartoon captions by~\citet{shahaf2015inside}. For our analysis, we construct the {\bf joke profile graph}, shown in Figure~\ref{fig:joke_profile}. It shows the proportion of time the editors substituted a word at each relative word position bin compared to if they randomly chose a word to substitute. Specifically, the red curve in the plot shows the proportion of replacement word locations if they were chosen randomly from those available in the editing task. The green curve shows the proportion of word locations actually chosen by our editors, and the blue curve shows the difference. We see that the blue line rises monotonically toward the end of the headline, meaning that editors tend to prefer replacing words later in the headline. The plot also shows the average funniness grade as a dotted line as a function of the position of the replacement word. It rises dramatically toward the end, showing that the funniest headlines were generally those with a replacement word toward the end.

\subsubsection{Superiority for Humor}
Jokes often express our feelings of superiority over someone else~\cite{sep-humor}. This can lead to frequent use of negative sentiment in jokes, as found by~\citet{mihalcea2007characterizing} in their analysis of humorous texts. We find similar support in our clusters of replacement words in Table~\ref{tab:nabil_clusters}, where the clusters labeled \emph{insults}, \emph{human deficiencies}, and \emph{corrupted} are all comprised of words that tend to denigrate other people, accounting for about 12\% of the substitute words inserted by our editors.

\section{Humorous Headline Detection}
In this section, we develop baseline classifiers to infer whether an edited headline is funny. 
Given our dataset, there are three possible combinations of information that we can use to detect humor: 
\begin{enumerate}
\itemsep-0.2em 
    \item Using only the edited headline to predict whether it is funny or not. 
    \item Using only the original headline to predict its potential for funniness.
    \item Using both original and edited headlines to jointly predict resulting funniness.
\end{enumerate}
We address the first of these scenarios. A classifier of this type could be used in a generate-and-test setting to create humorous headlines by trying different micro-edits.

To map the range of observed funniness grades (see Figure~\ref{fig:headline_funniness}) to the funny/not-funny classes, we sort our full dataset in decreasing order of mean funniness\footnote{We then sort by increasing standard deviations of grades to further rank headlines which tie on funniness, as lower standard deviation indicates stronger judge agreement.} scores, and we take the top $X\%$ of the data from each end, at size intervals of 10\%.
Note that each train/test split has an equal number of funny and not-funny headlines, establishing a 50\% majority class baseline for accuracy. 

We first trained a number of non-neural classifiers (logistic regression, random forest, SVM), using two feature sets: n-gram features (1, 2 and 3-grams combined) and features based on GloVe embeddings\footnote{This is our only feature set that uses the replaced word.} as shown in Figure~\ref{fig:gloveCorrelations}. We use 80\% of the data for training and 20\% for testing. We optimized hyperparameters for accuracy on 10-fold cross validation on the training set. The random forest classifier consistently performed best, so we only report its test set performance.

We also applied a neural baseline model using a single-layer bi-directional LSTM~\cite{hochreiter1997long} with 16 hidden units, a dropout of 0.5, and GloVe pre-trained embedding of the sequence of words in the edited headline. The training set was further split into 80\%-20\% splits for training and validation. We used a mini-batch size of 32 with up to 25 epochs to train our model, optimizing for cross-entropy.

Table~\ref{tab:classification_accuracy} shows the results obtained with our classifiers. LSTM performs better than random forest with either n-gram (Rf-ngram) or GloVe features (Rf-Glv), achieving our best accuracy of 68.54\% using $X$ = 10. We suspect that the reason for 
the LSTM's superior performance is that it learns predictive interactions between the semantics of the headline's words (via the GloVe embeddings) that trigger the humor\footnote{Using only words in the original headline produced accuracy in the mid-50\% range, suggesting that  the LSTM captures some humor impact of the replacement word as input and that some headlines have high potential for funniness.}.

Table~\ref{tab:classification_accuracy} also shows that accuracy generally decreases as $X$ increases, which is expected since higher $X$ implies a smaller separation between funny and not-funny classes, making classification harder. 
This is further corroborated by the observation that annotator-agreement scores (also shown in Table~\ref{tab:classification_accuracy}) decrease similarly with increasing $X$, 
indicating that funny and not-funny classes are easier to distinguish at the extreme ends of the dataset for both humans and machines alike.

\subsection{LSTM Classification Analysis}
We now investigate the test-set performance of the LSTM trained on the dataset obtained using $X$~=~$40$, the largest of our experimental datasets for which the class boundaries are distinct.

To analyze how well the classifier predicts the extremes in the test set, we obtained classification accuracy on distinct mean grade bins, presented in Table \ref{tab:partitioned_labels}. 
The LSTM is able to distinguish the far extremes ($\leq$0.4 and $>$1.6) of the test set much more convincingly than the headlines with mean grades in the interval (0.4,1.6]. We found a slightly negative correlation between classification accuracy and standard deviation of grades. 
Using additional judges for headlines with high standard deviation of grades would possibly improve annotator agreement and classification accuracy.

\begin{table}
    \centering
     \resizebox{\columnwidth}{!}{
    \begin{tabular}{|c||c|c|c|c|c|c|c|}
         \hline
         Bin & 0-.4    & .4-.8   & .8-1.2   & 1.2-1.6   & 1.6-2.0   & 2.0-2.4   &  2.4-3.0 \\ \hline
         Acc. & 71.5   & 61.1  & 52.0  & 61.0  & 68.6  & 68.3  & 76.2 \\ \hline
    \end{tabular}}
    \vspace{-6pt}
    \caption{LSTM accuracy for distinct grade bins (upper-bounds are inclusive) on the test set for $X=40$.}
    \label{tab:partitioned_labels}
        \vspace{-1pt}
\end{table}

The LSTM achieved a significantly lower accuracy when an entity (61.8\%) was replaced by the micro-edit compared to when a noun (64.5\%) or a verb (65.5\%) was replaced. For 5 of the 7 bins in Table \ref{tab:partitioned_labels}, the entity-replaced headline classification accuracy was lower than when the other two types were replaced, with the LSTM only achieving an accuracy of 47.9\% on the (0.8,1.2] bin for entity-replaced headlines. Although the classifier is never shown what has been replaced, it is better at assessing humor when the replaced word is not an entity. 
Our judges did have access to the replaced word, so we speculate this knowledge is important when the replaced word is an entity, especially when the entity triggers the judge's recollection of their world knowledge surrounding the entity, which the LSTM does not have. 
Another potential reason is that the pretrained GloVe vectors
are trained on web data (840 billion tokens obtained from Common Crawl) no more recent than 2014, which may not appropriately represent common entities in our 2017-2018 headline data.

Next, we qualitatively analyzed the LSTM's classification accuracy towards the two extremes of the dataset, some of which are shown in Table~\ref{tab:headline_edits}. 
Overall, the LSTM seems to suffer from a relatively high level of brittleness (possibly arising from the unusual writing style in headlines), 
where correct predictions could be obtained by very little modification to the text. For example, changing ``Trump $\rightarrow$ Trump's'' in ID 1 and deleting ``Essential Politics:'' in ID 3 fix their classification errors.
Quotes in headlines also confused the LSTM (e.g., ID 4) since it is sometimes non-trivial to discern the speaker of the quote in a headline.

The classifier often had difficulty figuring out humorous replacements that involve common-sense knowledge (e.g., IDs 7 and 8). Not surprisingly, it also failed to detect offensive replacements as in IDs 13 and 14, where the model probably recognized the incongruity and marked these as funny. World knowledge and cultural references were other challenges (e.g., IDs 4, 9 and 14).

The LSTM was able to figure out some of the obvious negative sentiments which were common in unfunny headlines (e.g., ID 15), and it detected some humor patterns resulting from using words that form a common (but funny in the context) n-gram with the adjacent words (e.g., IDs 5 and 6). 

Overall, our results show that there is a discernible signal separating funny and and not-funny headlines, even when using relatively shallow features that only take the content of the headline into account (modulo GloVe embeddings which are pretrained and hence contain semantic information gleaned from a larger corpus). We expect that further work, which could examine deeper relationships to current events, historical context, and common sense knowledge, will improve the ability to distinguish funny from not-funny beyond the baselines provided here.

\begin{table}
    \centering
         \resizebox{\columnwidth}{!}{
     \begin{tabular}{c|cc|c|ccc}
         $X$  & MaxUF    & MinF     & $\alpha$    & Rf-Glv   & Rf-ngram       & Lstm-Glv \\ \hline \hline
         10 & 0.2       & 1.8        & 0.66     & 60.27      & 65.56        & 68.54      \\ 
         20 & 0.4       & 1.4        & 0.49     & 61.67      & 63.41        & 67.21      \\ 
         30 & 0.6       & 1.2        & 0.37     & 59.43      & 62.96        & 66.11      \\ 
         40 & 0.8       & 1.0        & 0.27     & 57.45      & 59.35        & 64.07 \\ 
         50 & 0.8       & 0.8        & 0.20     & 55.63      & 56.52        & 60.63       
         
    \end{tabular}}
    \caption{Classification accuracy for various funniness-sorted dataset proportions and classifiers/feature sets. MaxUF is the highest score for the not-funny class, MinF is the lowest score for the funny class, and we also provide Krippendorff's $\alpha$ for judge agreement.} 
     \label{tab:classification_accuracy}
\end{table}

\section{Related Work}
Previous research on automated humor can be divided into work on datasets, analysis, detection, and generation. We will give examples of each.

Datasets are important for automated understanding of humor and for training models. Starting at the simplest linguistic level,~\citet{engelthaler2018humor} gathered almost 5,000 English words with funniness ratings for each one.~\citet{filatova2012irony} found 1,905 Amazon product reviews classified as either regular or ironic/sarcastic, and~\citet{khodak2017large} collected 1.3 million sarcastic statements from Reddit and a much larger set of non-sarcastic statements. \citet{mihalcea2005making} collected about 24,000 one-liner jokes, \citet{potash2017semeval} shared a dataset to rank funny tweets for certain hashtags, 
and \citet{miller2017semeval} created a task for pun detection.

Humor analysis, as we have done, is aimed at understanding what makes something funny. Building on the word-level corpus of~\citet{engelthaler2018humor},~\citet{westbury2018wriggly} developed models to predict the funniness of 4,997 words. Looking at multi-word, but still short text,~\citet{shahaf2015inside} analyzed cartoon captions in order to understand what made some funnier than others. The work that is most similar to ours is from~\citet{west2019reverse}, who looked at pairs of funny and normal headlines.
While we employed editors to create funny headlines from serious ones, they went the other way using a Web-based game, producing and analyzing 2,801 modified versions of 1,191 satirical headlines.

Humor detection is characterized by determining if a given text is funny or not. Examples include~\citet{khodak2017large}, detecting sarcasm in Reddit and
\citet{davidov2010semi} detecting sarcasm in Amazon product reviews and Twitter.~\citet{barbieri2014automatic} and ~\citet{reyes2012humor} showed how to detect humorous tweets, and~\citet{kiddon2011s} detected double entendres.

Generating humor is a difficult problem. Past work includes~\citet{binsted1997children} producing punning riddles, funny acronyms from~\citet{stock2003hahacronym}, jokes of the type ``I like my \underline{coffee} like I like my \underline{war}, \underline{cold}'' by~\citet{petrovic2013unsupervised}, and filling in Mad Libs\textsuperscript{\small{\textregistered}} by~\citet{hossain2017filling}. 
Our headline work has the potential to help in humor generation, moving away from jokes with a strong template to more free form.
\section{Conclusion and Future Work}
We have developed and released Humicroedit, a carefully curated dataset of 15,095 headlines with simple edits designed to make them funny. The dataset specifies the edits and also comes with five funniness scores for each edited headline. The simple replacement edits facilitate focused analysis on what causes the humor. We showed how our data supports, in a quantitative way, humor theories about length of joke, incongruity, superiority, and setup/punchline. Finally, we developed baseline classifiers that show how well we can distinguish funny edits from non-funny edits using simple linguistic features.

We expect our dataset will facilitate research in humor and natural language processing. Headlines present unique challenges and opportunities, because their humor is largely topical, depending on a knowledge of current events and prominent people and entities.

Future work with this data could include deeper features for assessing humor. We expect that humor detection would likely improve using features that incorporate world knowledge and common sense. Likewise, there may be something to learn by analyzing topical jokes from professional comedians. With our single-word edits, this analysis becomes easier, because we are looking at the minimal change in a headline to make it funny. Additionally, if we can better understand what makes a headline funny, we may be able to automatically generate funny headlines and even personalize them to particular readers.

\section*{Acknowledgments}
We thank Daniel Gildea for carefully reviewing our paper and for his advice on the machine learning experiments. We also thank the NAACL reviewers for their various helpful suggestions. 

\bibliography{headlineHumor}
\bibliographystyle{acl_natbib}

\end{document}